\documentclass[conference]{IEEEtran}
\IEEEoverridecommandlockouts
\usepackage{cite}
\usepackage{amsmath,amssymb,amsfonts}
\usepackage{algorithmic}
\usepackage{graphicx}
\usepackage{textcomp}
\usepackage{subcaption}
\usepackage{xcolor}
\usepackage{flushend} 
\usepackage{wrapfig}
\usepackage{multirow}
\usepackage{hyperref}

\definecolor{red}{rgb}{1.00,0.00,0.00}
\definecolor{blue}{rgb}{0.00,0.00,1.00}

\newcommand{\cblue}[1] {\textcolor{blue}{#1}}

\def\BibTeX{{\rm B\kern-.05em{\sc i\kern-.025em b}\kern-.08em
    T\kern-.1667em\lower.7ex\hbox{E}\kern-.125emX}}
\begin{document}

\title{{\small \vspace{-1.5cm} \href{https://humanoids-2020.org/}{Accepted in IEEE-RAS International Conference on Humanoid Robots (Humanoids2020)}}\\ \vspace{1.5cm}Open-Ended Fine-Grained 3D Object Categorization by Combining Shape and Texture Features in Multiple Colorspaces\\}

\author{\IEEEauthorblockN{Nils Keunecke, S. Hamidreza Kasaei}
\IEEEauthorblockA{\textit{Department of Artificial Intelligence, University of Groningen, Groningen, Netherlands} \\
n.keunecke@student.rug.nl, hamidreza.kasaei@rug.nl}
\thanks{We thank NVIDIA Corporation for their generous donation of GPUs which was partially used in this research.}
}

\maketitle

\begin{abstract}
As a consequence of an ever-increasing number of service robots, there is a growing demand for highly accurate real-time 3D object recognition. Considering the expansion of robot applications in more complex and dynamic environments, it is evident that it is not possible to pre-program all object categories and anticipate all exceptions in advance. Therefore, robots should have the functionality to learn about new object categories in an open-ended fashion while working in the environment. Towards this goal, we propose a deep transfer learning approach to generate a scale- and pose-invariant object representation by considering shape and texture information in multiple color spaces. The obtained global object representation is then fed to an instance-based object category learning and recognition, where a non-expert human user exists in the learning loop and can interactively guide the process of experience acquisition by teaching new object categories, or by correcting insufficient or erroneous categories. In this work, shape information encodes the
common patterns of all categories, while texture information is used to describes the appearance of each instance in detail. Multiple color space combinations and network architectures are evaluated to find the most descriptive system. Experimental results showed that the proposed network architecture outperformed the selected state-of-the-art approaches in terms of object classification accuracy and scalability. Furthermore, we performed a real robot experiment in the context of \textit{serve\_a\_beer} scenario to show the real-time performance of the proposed approach.
\end{abstract}


\section{Introduction}
At the end of this decade, autonomous mobile robots are believed to be used in our life as service robots, self-driving cars, and collaborative industrial robots. However, as the environment becomes more complex, it is important that the robots use an accurate and robust perception system. To work in such a dynamic environment and safely interact with human users, robots need to know which kind of objects exist in the scene and where they are. Therefore, object detection and recognition play important role in collaborative robots. Three-Dimensional (3D) Object recognition is a fundamental research problem in computer vision and robotic communities. Although recent approaches have shown promising results under structure environments, there are several unresolved issues. In particular, most recent approaches follow the \texttt{\small train-then-test protocol}, which means that the robot is trained once all
data has been gathered. Therefore, the performance of the robot strongly dependent on the quality and quantity of training data. Furthermore, the knowledge of such robots is fixed after the training phase, and any changes in the environment require complicated, time-consuming, and expensive robot re-programming by expert users.

\begin{figure}[!t]
\centering
\includegraphics[width=\linewidth]{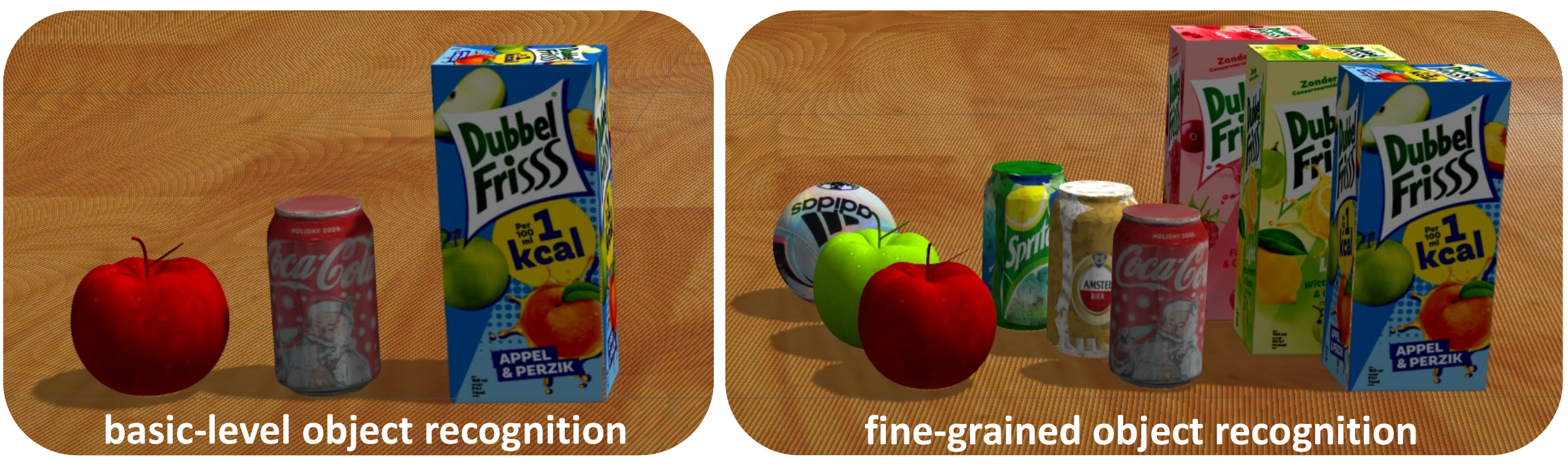}
\caption{\small Basic-level object categorization such as differentiating the left three objects has been done for years and high performance is achieved frequently. In contrast, differentiating between object classes in fine-grained object categorization tasks like the three sets of three objects on the right side is difficult if either only shape or color information is used. This paper shows how a combination of texture and shape significantly increases the performance in such tasks.}
\end{figure}
A dynamic environment makes it impossible to pre-program all possible object categories, and anticipate all exceptions before the robot goes into operation. Therefore, the robot should have the ability to learn about new object categories in an online and open-ended fashion. In order to meet these requirements, a robot should be able to update the model of existing categories as new instances are encountered and create a new model once facing a new object category. While this procedure can be partly supervised in the form of human-in-the-loop feedback (i.e., non-expert human users can interactively guide the process of experience acquisition by teaching new categories or by correcting insufficient or erroneous categories), the robot also has to learn independently from on-site experiences in its environment. In human-centric environment, the robot frequently encounters a new object that can be either not similar (e.g., \textit{apple} vs. \textit{juice box}) or very similar (e.g., \textit{coke can} vs. \textit{beer can}) to the previously learned object categories in terms of texture and shape. This is a very challenging task to learn about fine-grained object categories using a few examples since deep learning approaches often need large-scale training data to avoid over-fitting. Furthermore, describing objects only by either shape or color will likely lead to confusion eventually. In such scenarios, the object representation plays a significant role since it will use in both training and testing phases.

\begin{figure}[!t]
\centering
\includegraphics[width=\linewidth]{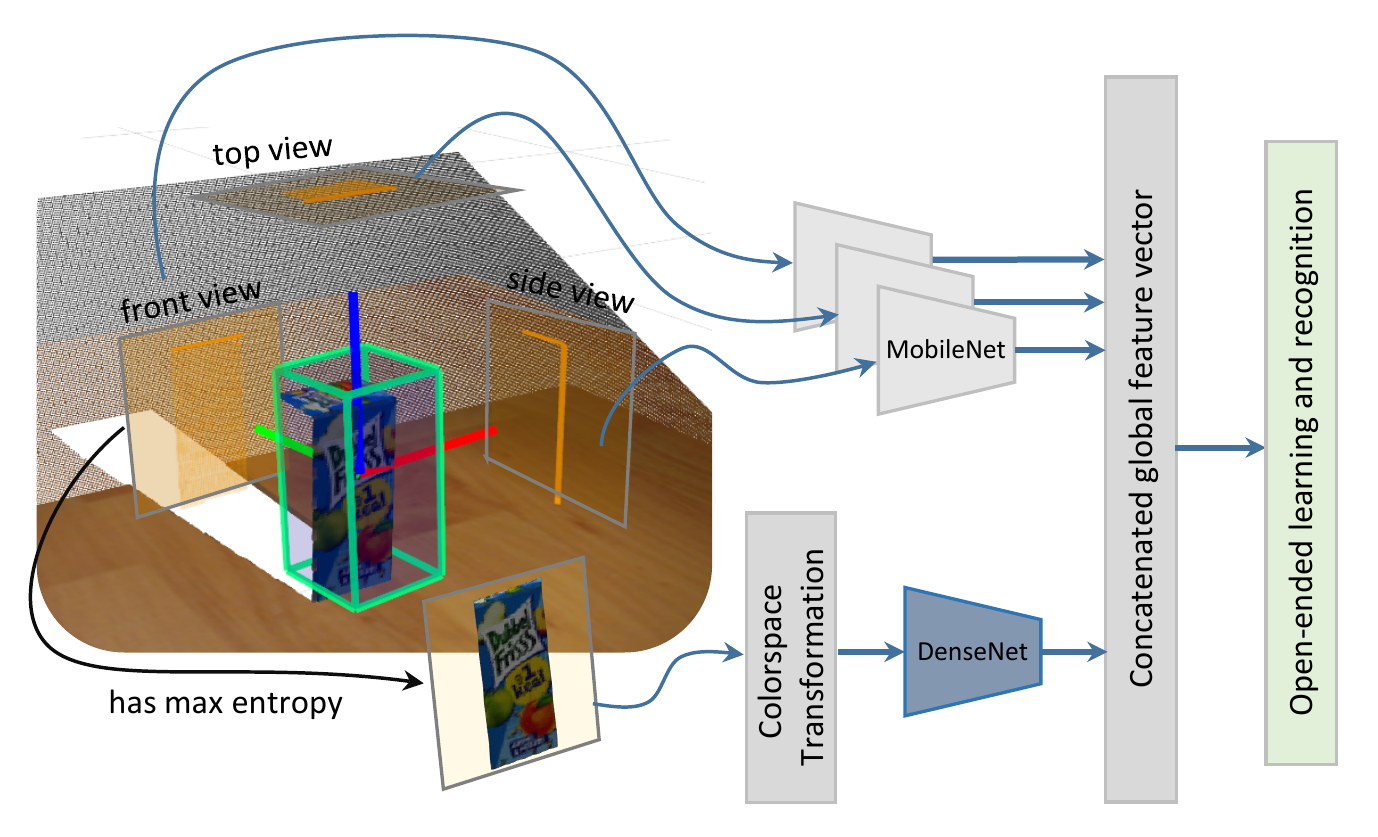}
\caption{\small The overall system architecture: in the first step the point cloud of the object is processed to generate three depth orthographic projections for encoding the shape of the object, and one colored projection is rendered from the view that has maximum entropy. On the top-right of the figure, the shape information is processed via MobileNetV2. On the bottom-right, the RGB color information is transformed into different color spaces and evaluated individually either via a DenseNet or the MobileNetv2, depending on the setup. The output of both sides is concatenated into a global feature vector which is finally used for learning and recognition purposes.}
\label{fig:overal}
\end{figure}

In this work, we propose an approach to represent an object based on depth and colored orthographic (top, side, and front) views.
In particular, we use three depth views of an object to represent the geometrical properties (basic-level), and a colored orthographic view, i.e., rendered from the direction that has maximum entropy, is used for describing the texture of the object (fine-grained). Towards this goal, we first construct a unique Local Reference Frame (LRF) for the object, and then, render orthographic projections by exploiting the LRF. Each projected view is then fed to a CNN to obtain a view-wise deep feature. The representation of the object is finally constructed by concatenating the color and depth representations.  It is worth mentioning that the final representation is a pose- and scale-invariant, and designed with the objective of supporting accurate 3D object recognition. The required steps leading to the eventual deep object representation for a \textit{juice box} object are shown in Fig.~\ref{fig:overal}. 

The remainder of this paper is organized as follows: first, related work will be presented. Section~\ref{sec:method} will discuss the architecture of the system presented in this paper. In section~\ref{sec:results} the results of the offline and online evaluation are presented. This section additionally features the robot demonstration. Finally, in section~\ref{sec:conclusion} conclusions will be drawn and further work is proposed.

\section{Related Work}
\label{sec:related_work}


Three-dimensional object recognition has become a field of fundamental importance in computer vision, pattern recognition, and robotics. Convolutional Neural Networks (CNNs) are frequently used for image classification purposes\cite{al2017review}. Recently a tendency towards deeper network architectures can be observed\cite{simonyan2014very}\cite{zhang2019deeper}. While excellent results have been reported, it has become evident that due to the enormous number of parameters, computation time and memory requirements have to be considered as well. Once in operation, the image classification algorithm may have limited resources, for example in robot applications. Generally, there is a trend towards optimizing the descriptiveness to computational complexity ratio of network architecture\cite{ColorNET}\cite{cheng2017survey}.

Additionally, it has been identified that it is impossible to pre-train neural networks for 3D object recognition entirely. Open-Ended learning approaches have recently become more popular \cite{lucas1995towards}\cite{kasaei2015interactive}\cite{oliveira2015concurrent}. These approaches address the problem of CNNs to adapt to an increasing number of categories as this would require reshaping of the typology of the network. CNNs tend to require a lot of training data to perform accurately, which is unfeasible in open-ended learning as new categories have to be learned opportunistically with very few instances in the beginning, and incrementally updated as more instances are presented. Among others, Kasaei et al.\cite{orthographicNET} reported that their transfer learning approach yields excellent results for such tasks. These open-ended network architectures are usually pre-trained on the imageNet dataset\cite{imagenet}.

\begin{figure}[!b]
\centering
\includegraphics[width=\linewidth]{img/image_2020_09_03T11_36_45_439Z.png}
\caption{\small Representing object based on shape-only or color-only is not enough to distinguish very similar objects. In contrast, combination of texture and shape features makes it possible to robustly recognize very similar objects~\cite{Ayoobi2020LocalHDPI}.}
\label{fig:similar_objects}
\end{figure}

OrthographicNet~\cite{orthographicNET} is one of several recently proposed view-based approaches \cite{view1}\cite{kanezaki2018rotationnet}\cite{mvcnn}, which tend to show superior performance compared to volume-based~\cite{volume1}\cite{volume2} and point-based  approaches  \cite{pointnet}\cite{pointnet++}. View-based approaches construct 2D representations based on the point cloud of the object. In contrast, volume-based approaches use the point cloud to construct a 3D voxel-grid. Point-based approaches directly operate on the point cloud, obtained from the object. All approaches use the obtained representation as to their respective input for a neural network (typically some form of CNN). In contrast to other view-based approaches, the OrthographicNet uses a partial point cloud of the object to generate 2D orthographic projections of the object. For robotics applications and other real-world implementations, multi-view representations of objects are problematic due to the lack of scenarios where objects are fully observable. Unlike our approach, all these approaches only considered shape information and discard color information completely.

In general, most object recognition algorithms can be divided in two groups with either focus on shape-information~\cite{belongie2002shape}\cite{kazmi2013survey}\cite{good} or on color-information~\cite{krizhevsky2012imagenet}\cite{color2}. While shape-only approaches frequently struggle with similarly shaped objects \cite{investigating}~(Fig. \ref{fig:similar_objects}), color-based approaches are volatile to shadows and illumination \cite{poth2016breaking} and tend to have a bias towards texture \cite{geirhos2018imagenet}.

Even though there exist several state-of-the-art shape-only approaches, the neuroscientific argument has been made by Bramao et al., that for humans color-information is essential for object recognition~\cite{ColorOR}. There are recent findings that suggest that neural networks can profit as well if they combine the available shape information with color information~\cite{investigating}\cite{tsai2018simultaneous}\cite{zia2017rgb}. Several strategies to achieve this interaction exist. Three approaches are presented in order of increasing computational complexity. \textit{(i)} a color constancy value can be calculated to find the average color of an object \cite{investigating}. While this approach already increases the performance, it lacks the ability to detect complicated textures. \textit{(ii)} Shape information can be evaluated in parallel to an RGB image of the object \cite{asif2017rgb} \cite{cruz2013object}. This reduces in particular the overall tendency of the purely color-based descriptors to be biased towards texture and will help shape-only descriptors to differentiate between objects of similar shape (like two different soda cans).  \textit{(iii)} Gowda et al.~\cite{ColorNET} have found that combining different color spaces by transforming the RGB image improves the reported overall accuracy as different features of an object are represented differently in different color spaces. The third approach combines color information with shape information. It searches for the most optimal combination of color spaces and combines them with a state-of-the-art shape descriptor, such as the aforementioned OrthographicNet. It is expected that the resulting architecture has all the advantages of shape and color descriptors while the two parts of the network architecture mitigate each other's weaknesses.


\section{Method}
\label{sec:method}
In this work, we assume that objects are placed on a surface, e.g., table, and already segmented from the input point cloud. Each object is represented by a set of points, $p_i \in \{1, ..., n\}$, where each point contains standard RGB color information [R, G, B], as well as a depth value in three axes [x, y, z].

\begin{wrapfigure}{r}{0.3\linewidth}
\vspace{-2mm}
  \begin{center}
    \includegraphics[width=0.9\linewidth]{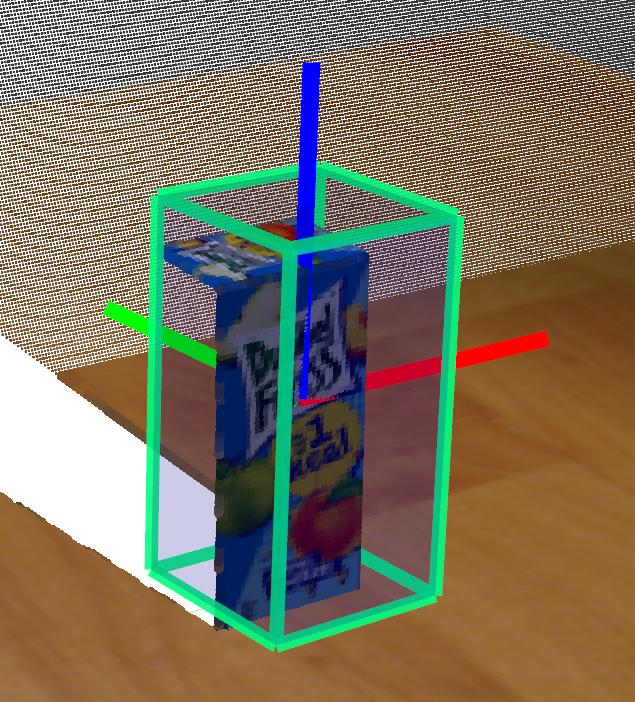}
  \end{center}
  \caption{\small An example of object reference frame construction and bounding box estimation for a juice box. The red, green, and blue lines show the \textbf{X}, \textbf{Y}, and \textbf{Z} axes, respectively.}
  \label{fig:rf}
  \vspace{-3mm}
\end{wrapfigure} 
For rendering orthographic images of a 3D object, we put three virtual cameras around the object: where the $\textbf{Z}$ axes of a camera is parallel with one of the principal axes of the object, and pointing towards the centroid of the object, and perpendicular to plane constructed by the other two axes of the object. Therefore, it is necessary to first create a Local Reference Frame (LRF) for the object. Towards this goal, we first compute the geometric center of the object, which is defined as the arithmetic mean position of all the points of the object. Afterwards, we construct a LRF by performing eigenvalue decomposition analysis on the normalized covariance matrix, $\Sigma$, of the object, i.e., $\Sigma\textbf{V}=\textbf{EV}$, where $\textbf{E} = [e_1, e_2, e_3]$ contains the descending sorted eigenvalues, and $\textbf{V} = [\vec{v}_1, \vec{v}_2, \vec{v}_3]$ shows the eigenvectors. In this work, the largest eigenvectors, $\vec{v}_1$ is considered as $\textbf{X}$ axis. Since we assumed the object is laying on a table, the $\textbf{Z}$ axis of the object is set to the direction that is perpendicular to the table (gravity direction). We finally define the $\textbf{Y}$ axis as the cross product of $\textbf{X} \times \textbf{Z}$. The object is then transformed to be placed in the reference frame (see Fig.~\ref{fig:rf}). Finally, from each camera pose, we project the point cloud of the object into a depth image using the z-buffering and orthogonal projection methods~\cite{liu2019soft}.

After rendering orthographic projects, the depth and color views are separated into two input streams, one for color information and one for shape information (see Fig. ~\ref{fig:overal}). In particular, three scale and rotation invariant projections are rendered based on the z-buffering technique to encode geometrical information of the object. The obtained orthographic depth projections, namely the \textit{front-}, \textit{side-}, and \textit{top-view} are fed into the OrthographicNet~\cite{orthographicNET}. In this work, the MobileNetv2~\cite{sandler2018mobilenetv2} was used for each individual projection, as Kasaei et al. have reported the best results with this network architecture~\cite{orthographicNET}.

\begin{figure}[!t]
\centering
\begin{subfigure}[t]{0.3\linewidth}
\includegraphics[width=0.99\linewidth]{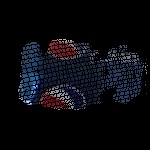}
\end{subfigure}
\begin{subfigure}[t]{0.3\linewidth}
\includegraphics[width=0.99\linewidth]{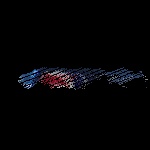}
\end{subfigure}
\begin{subfigure}[t]{0.3\linewidth}
\includegraphics[width=0.99\linewidth]{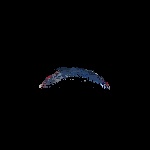}
\end{subfigure}
\caption{\small An illustrative example of rendering three colored orthographic projections from the partial point cloud of a soda can.}
\label{fig:soda_can}
\end{figure}

To process the texture information, one intuitive option is to rendered three colored orthographic projections of the object (similar to orthographicNet). However, a set of preliminary tests showed that no performance improvements could be gained using all three projections but rather that a single projection performed significantly better than the other two (see Fig.~\ref{fig:soda_can}). This is related to the aforementioned texture-bias of color-based classifiers. We use viewpoint entropy to define which of the three orthographic projections is best for further processing (see Fig.~\ref{fig:overal}). The underlying reason for considering viewpoint entropy as the metric of selecting the best view is that viewpoint entropy nicely takes into account both the number of occupied pixels and their values.  Entropy is defined as a metric to measure how much information is contained in a single projection, $\textbf{X}$, and can be calculated as: 

\begin{equation}
     H(\textbf{X}) = -\sum_{i=1}^{n} p(x_i)~log_b p(x_i)
    \label{equation1}
\end{equation}

All three orthographic RGB projections of an object are compared and the projection with the highest entropy (e.g., most information) is selected for further processing. Using only one projection instead of three effectively reduces the network size of the color evaluation by two-thirds which will be beneficial for the computational performance. Based on the work of Gowda and Yuan \cite{ColorNET}, the following color spaces were selected: RGB, HED, HSV, LAB, YCbCr, YIQ, and YUV. Additionally, grayscale was included for a total of 8 color spaces. While all color spaces represent the same color, they use different mathematical models to represent that color. The resulting numerical differences have an impact on neural networks where different filters are learned depending on the color space. Input images are transformed into each color space using the respective transformation as displayed exemplary in the following equation, shown for a color transformation from RGB to the YUV color space.
\begin{equation}
\scriptsize
\begin{bmatrix}
Y\\ 
U\\ 
V
\end{bmatrix}
= 
\begin{bmatrix}
0.299 & 0.587 & 0.114\\ 
-0.168 & -0.331 & 0.500\\ 
0.500 & -0.418 & -0.0813 
\end{bmatrix}
\begin{bmatrix}
R\\ 
G\\ 
B
\end{bmatrix}
+
\begin{bmatrix}
0\\ 
128\\ 
12
\end{bmatrix}
\end{equation}
After the color space transformations, input images are fed into the respective network to encode the texture information of the object.  


\begin{wraptable}{r}{0.6\linewidth}
\vspace{-3mm}
\caption{\small Properties of the used CNNs.}
\resizebox{\linewidth}{!}{
    \begin{tabular}{|c|c|c|}
    \hline
    Model          & DenseNet & MobileNetv2 \\\hline\hline
    Depth          & 40                & 88          \\\hline
    Feature length & 132 float         & 1280 float \\\hline
    Input size     & 64 x 64           & 224 x 224   \\\hline
    Parameters     & 0.225M            & 2.25M       \\\hline
    Size           & 3 MB              & 14.5 MB     \\\hline    
    \end{tabular}
    \label{tab:table1}}
\end{wraptable}
In this paper two neural network architectures were used to evaluate color information. A DenseNet 40-12-BC~\cite{huang2017densely} is used, which is 40 layers deep and has a growth factor of $12$. The BC refers to compression layers at the end of each dense block. Additionally, the MobileNetv2 (here refer to as MobileNet) was used, which is significantly deeper than the DenseNet and has almost 10 times the parameters as the DenseNet (Table~\ref{tab:table1}). Each color space was trained on both networks using Washington RGB-D dataset~\cite{washington-d} (refer to Section~\ref{sec:offline} for more details). 

To classify the learned representations, a MLP consisting of two linear layers with output size $1024$ and $256$ respectively were used. Both layers use rectified linear units (ReLU) and a dropout layer with rate $0.2$. Finally, predictions are made by a linear layer with softmax activation. Statistic Gradient Descent (SGD)\cite{SGD} is used for optimization with a momentum of 0.9 and nesterov momentum. The initial learning rate is set to $lr_0 =0.05$ and an exponential learning rate decay $lr = lr_0 * e^{-kt}$ with $k=-0.02$, where t is the number of iterations. No warm-up nor restarts were used. This configuration was used for all offline evaluations. Training was carried out for $150$ epochs with a batch size of $128$. In addition, a global weight decay of $10^{-6}$ was used.


\section{Results}
\label{sec:results}

To evaluate the proposed approach, a total of three experiments were performed: the offline-evaluation, online-evaluation, and a real-time robot demonstration. All tests were performed with a PC running Ubuntu $18.04$ with a $3.20$ GHz Intel Xeon(R) i$7$ CPU, and a Quadro P$5000$ NVIDIA.
In the case of offline and online evaluations, experiments were carried out using the Washington RGB-D dataset~\cite{washington-d}. This dataset contains 300 common household items, which are organized into 51 classes. From the available 250000 views, 50000 orthographic projections were generated and divided into a train, validation, and test set with a 70/15/15 split. During the offline evaluation, Average Class Accuracy (ACA) over 10 trials is reported.

\subsection{Offline Evaluation}
\label{sec:offline}

\subsubsection{Color space evaluation}

Firstly, the individual color spaces were evaluated. From the results reported  in Table ~\ref{tab:table2}, it can be observed that on average the MobileNet performed about 2\% better than the DenseNet. However, in the HSV and the YCbCr color spaces, the DenseNet showed better results than the MobileNet. The best average class accuracy (ACA) was $98.56\%$, which was obtained by MobileNet architecture and RGB color space configuration. In the case of DenseNet, the best ACA, obtained with the YCbCr color space, was $97.44\%$. Notably, some color spaces perform significantly worse in this classification task (i.e., Grayscale, YIQ, and HED).
\begin{wraptable}{r}{0.6\linewidth}
\caption{\small Object recognition accuracy for all color spaces.}
\resizebox{\linewidth}{!}{%
\begin{tabular}{|c|c|c|}
\hline
Colorspace & DenseNet  & MobileNet \\\hline\hline
RGB        & 96.56\%           & \textbf{98.56\%}     \\\hline
HED        & 93.59\%           & 96.86\%     \\\hline
HSV        & 97.32\%           & 96.40\%     \\\hline
LAB        & 96.37\%           & 96.87\%     \\\hline
YCbCr      & \textbf{97.44\%}  & 97.19\%     \\\hline
YIQ        & 92.33\%           & 92.51\%     \\\hline
YUV        & 95.24\%           & 97.43\%     \\\hline
Grayscale  & 91.55\%           & 95.80\%    \\\hline
\end{tabular}
}
\label{tab:table2}
\end{wraptable}
It is evident that not all color spaces may be beneficial to improve object recognition accuracy. As the network architecture both perform well, though on different color spaces, the combination of color spaces was optimized on both color spaces. 

\subsubsection{Color space optimization}
In this round of experiments, we optimize the color spaces for finding the combination that yielded the highest average class accuracy. Due to the different sizes of the feature vectors between the DenseNet and the MobileNet, color space optimization was carried out architecture specific. It was found that the best combination of color spaces for the MobileNet as well as the DenseNet was a combination of the RGB, HSV, YCbCr, and YUV color space at $98.84\%$ and $98.16\%$, respectively. As computational performance is a key metric of evaluation and additional color spaces obviously impact increase the computation steps, Table \ref{tab:table4} summarizes the best combinations of one to four colorspaces for both network architectures. Neither benefits from more than four colorspaces.

\begin{table}[!b]
\centering
\caption{\small Colorspace optimization for DenseNet and MobileNet.}
\resizebox{0.9\linewidth}{!}{%
\begin{tabular}{|c|l|c|}
\hline
Model & Colorspace combination & ACA (\%) \\\hline\hline
\multirow{ 4}{*}{DenseNet} &YCbCr   & 97.44\%  \\\cline{2-3}
& YCbCr, HSV        & 97.48  \\\cline{2-3}
& YCbCr, HSV, RGB         & 97.67  \\\cline{2-3}
& YCbCr, HSV, RGB, YUV    & \textbf{98.12}  \\\hline\hline

\multirow{ 4}{*}{MobileNet} & RGB & 98.56  \\\cline{2-3}
& RGB, YCbCr                      & 98.69  \\\cline{2-3}
& RGB, YCbCr, YUV                 & 98.79  \\\cline{2-3}
& RGB, HSV, YCbCr, YUV            & \textbf{98.89} \\\hline
\end{tabular}
}
\label{tab:table4}
\end{table}
Once the best color space combination is obtained, the final and complete system architecture can be constructed. In particular, an object representation is finally constructed by concatenating the two individual shape and color feature vectors of the object. It should be noted that maximum or average pooling cannot be applied here. The underlying reason is that to encode the geometrical properties of the object, all orthographic projections are used while for representing the texture of the object, only the orthographic view of the object with the maximum entropy is used.
\subsubsection{Evaluation of the final network architecture}
\label{sec:results_color_shape}
The Washington RGB-D dataset is known to be color biased~\cite{investigating} and because of that in the previous round of experiments, all evaluations consistently yielded higher average class accuracies (ACA) than the best accuracy (i.e., $86.85\%$)reported for the OrthographicNet~\cite{orthographicNET} which only evaluates shape information. To investigate the optimal balance of color and shape information, a color weight vector $w$ is introduced and applied to the color feature vector. Similarly, a $(1-w)$ weight vector is applied to the shape feature vector:
\begin{equation}
    \textbf{f}(x,y,z,h) = (1-w) * \textbf{f}_s(x,y,z) + w * \textbf{f}_c(h)
\end{equation}
\noindent where $\textbf{f}_s$ and $\textbf{f}_c$ stand for shape and color feature vectors, respectively. The combined feature vector $\textbf{f}(x,y,z,h)$ is obtained by weighting the shape and the color feature vectors, and $x, y, z$ are the views of the respective orthographic projections, and $h$ is the color image of the view with the maximum entropy.
In this round of experiments, to reduce the overhead of the color space transformation, only for this part of the evaluation, all images were transformed beforehand. The training was carried out on the Washington-D dataset following the protocol laid down in Section 4. The ACA over $10$ trials is summarized in Table~\ref{tab:table5}. 

\begin{wraptable}{r}{0.6\linewidth}
\vspace{-1mm}
\centering
\caption{\small Average class accuracy for the combined network of color and shape information.} 
\resizebox{0.9\linewidth}{!}{%
\begin{tabular}{|c|c|c|}
\hline
weight & DenseNet (\%) & MobileNet (\%) \\\hline\hline
0      & 90.56            & 90.56     \\\hline
0.2    & 97.44            & 97.51     \\\hline
0.4    & 98.10            & 98.48     \\\hline
0.6    & \textbf{99.14}   & 99.00     \\\hline
0.8    & 99.00            & \textbf{99.07}     \\\hline
1.0    & 98.12            & 98.89    \\\hline
\end{tabular}}
\label{tab:table5}
\end{wraptable}
Based on the results reported in ~\ref{tab:table5}, it can be concluded that the model with DenseNet ($w = 0.6$) achieved the highest ACA ($99.14\%$) and the MobileNet with $w$ to $0.8$ obtained the second-best result. It should be noted that the MobileNet has $\approx9.75M$ parameters while the DenseNet has only $\approx0.75M$ parameters. By comparing all results, it is visible that the combination of color and shape outperformed both shape-only ($w=0.0$) and color-only ($w=1.0$) settings. While the classifier is theoretically able to learn the optimal weight between color and shape feature vectors during training, minor improvements can be observed from the weight initialization. Furthermore, the optimal weight between color and shape information will be used in the next step of the evaluation, where the open-ended capabilities of the network architecture were tested. 

\subsection{Online Evaluation}

 To evaluate the scalability of the proposed approach in the open-ended scenario, we used a recently introduced \texttt{test-then-train} protocol~\cite{investigating}. We developed a simulated user to interact with the robot using three different actions, including \textit{teach}, \textit{ask}, and \textit{correct}. The simulated teacher should be connected to a large object dataset, therefore, the Washington RGB-D dataset \cite{washington-d} is used in this round of evaluation. The main idea is to let the robot learns a new object category with the help of a simulated teacher (human-in-the-learning-loop).  Towards this goal, we replaced the classification part of the network with an instance-based learning approach \cite{kasaei2015interactive}. More specifically, an instance-based learning and recognition (IBL) approach considers category learning as a process of learning about the instances of a category, i.e., a category is represented by a set of known instances, $C \Longleftarrow\{\textbf{f}_1,\textbf{f}_2,\dots,\textbf{f}_n\}$, where $\textbf{f}_i$ is the representation of an object, as discussed in section~\ref{sec:results_color_shape}. It should be noted that IBL is a baseline approach to evaluate object representations. An advantage of the IBL approaches over other machine learning methods is the ability to rapidly adapt an object category model to a previously unseen instance by storing the new instance or by throwing away an old instance.

The teacher first teaches two new object categories to the robot using the \textit{teach} action. In the case of \textit{``teach''}, the learning agent stores the object views in its perceptual memory. In the case of \textbf{ask}-action, the teacher selects a previously unseen view of a known category and asks the robot to predict the category label of the object. If misclassification happens, the simulated teacher uses the \textbf{correct}-action to correct the system and sends the true category of the object, and the agent updates the respective category model using the misclassified object. 
The simulated teacher sequentially picks unseen objects from the known categories and asks the robot to recognize them. The teacher progressively estimates the recognition accuracy of the agent using a sliding window and a new category is introduced, when all known categories tested at least once and the recognition accuracy exceeds the protocol threshold ($\tau = 0.67$, meaning accuracy is at least twice the error rate). Should the system fail to reach this threshold after $100$ iterations, the experiment is aborted by the simulated user, as it can be concluded that the robot no longer has the capability to learn additional categories. Following this protocol, it is possible to simulate an environment in which the robot is simultaneously learning and recognizing. Since the sequence in which objects are introduced influences the resulting accuracy, we conducted $10$ experiments and reported the average of metrics.

\begin{table}[!b]
\centering
\caption{\small Summary of the online evaluation using RGB-D dataset.}
\resizebox{0.4875\textwidth}{!}{%
\begin{tabular}{|c|c|c|c|c|c|}
\hline
Method & QCI     & NLC   & AIC   & GCA  & APA  \\\hline\hline
RACE  \cite{oliveira20163d}                & 382.10  & 19.90 & 8.88  & 0.67 & 0.78 \\\hline
BoW \cite{bow}              & 411.80  & 21.80 & 8.20  & 0.71 & 0.82 \\\hline
Local-LDA \cite{local-lda}    & \textbf{262.60}  & 14.40 & 9.14  & 0.66 & 0.80 \\\hline
GOOD \cite{good}                 & 1659.20 & 39.20 & 17.28 & 0.66 & 0.74 \\ \hline
OrthographicNet(*)     & 1342.60 & \textbf{51.00} & 8.97  & 0.77 & 0.80 \\\hline\hline
Our + DenseNet (*) & 1409.10 & \textbf{51.00} & 10.28 & 0.75 & 0.77 \\\hline
Our + MobileNe(*)       & 1329.10 & \textbf{51.00} & \textbf{7.97} & \textbf{0.81} & \textbf{0.83}\\\hline
\multicolumn{6}{c}{ \centering (*) indicates that the stopping condition was ``\textit{lack of data}''.}
\end{tabular}}
\label{tab:table6}
\end{table}

At the beginning of the evaluation, the robot does not know about any categories but is pre-trained on the Washington RGB-D dataset.  By continuously introducing new instances of a category to the robot, the model of the category improves over time. As the dataset is limited in terms of categories, it may be possible that a robot was able to learn all available categories. This is indicated with a \textit{lack of data} in Table \ref{tab:table6}. The performance of the online evaluation highly depends on the order in which categories and views are selected by the simulated user. To account for this factor, all experiments in this section were carried out 10 times. For better comparability and more precise estimation of the potential performance of this work, five evaluation metrics are used. QCI denotes the number of question-correct iterations, that was necessary to learn the categories. This acts as a measure of how fast the robot learned. NLC is the average number of all categories learned by the robot. AIC is the average number of instances per category. Finally, the global class accuracy (GCA) and the Average Protocol Accuracy (APA) indicate how well the robot performs. It is worth mentioning that the ability to provide `real-time' 3D object recognition can not be abandoned due to its importance in applications in service robots, and collaborative robots. Therefore, memory usage (NLC) and computation time (\#QCI) have to be used as performance metrics during the evaluation of any robot perception system as well.

The obtained results for these experiments in relation to other recent approaches in open-ended object recognition such as BoW, RACE, Open-Ended LDA, and GOOD are reported in Table \ref{tab:table6}. It can be observed that our approach with MobileNet performed the best in all metrics. Particularly, not only it learns the fastest with $1329.10$ question-correction iterations but the average instances per category decreased by 1 compared to the shape-only OrthopgraphicNet. Additionally, the network architecture shows a $3-4\%$ performance increase in the GCA and APA evaluation metrics. The excellent scalability of the OrthographicNet can still be observed as all 10 experiments for both new network architectures have consistently learned all 51 categories.

However, the significant performance improvement from a purely shape-based descriptor to a robot that combines shape and color information as seen in the offline evaluation was not nearly as significant in the online evaluation. The network architecture which used the DenseNet even performed slightly worse than the original OrthographicNet. The DenseNet model learned all existing categories a bit slower than OrthographciNet and the proposed approach with MobileNet (i.e., $1409.10$ iterations vs. $1342.60$, and $1329.10$ iterations for OrthographciNet and Our approach with MobileNet, respectively). This pattern repeated for GCA and APA metrics. A possible cause for the slightly lower performance of the DenseNet is the comparatively shallow depth of the network and the small feature vector generated by the DenseNet. 

\subsection{Real-time robot demonstrations}

\begin{figure}[!b]
\centering
\includegraphics[width=0.95\linewidth, trim= 0cm 0.1cm 0cm 0cm,clip=true]{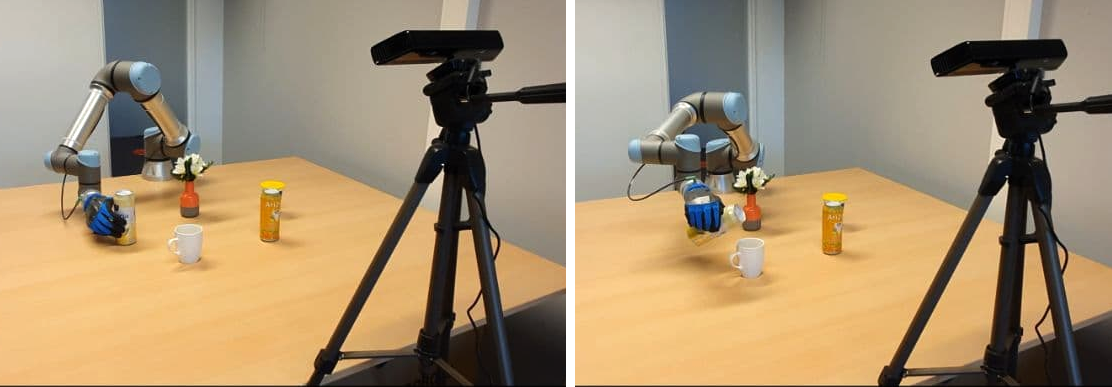}
\caption{\small Our experimental setup for the real-robot experiment consists of a table, a Kinect sensor, and a UR5e robotic-arm.}
\label{fig:setup}
\end{figure}

\begin{figure*}[!t]
\centering
\includegraphics[width=\linewidth]{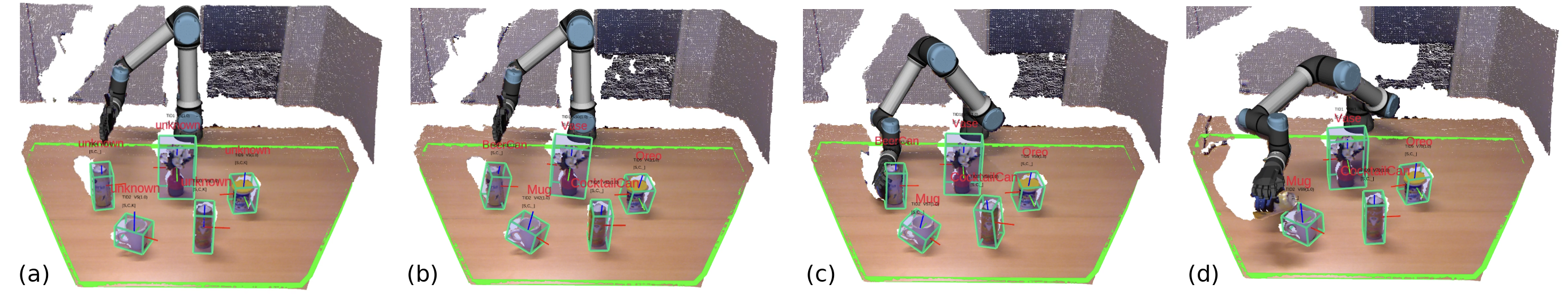}
\caption{\small The performance of the robot during the \textit{serve\_a\_beer} scenario: (\textit{a}) the table and all table-top objects are detected as indicated by the green bounding boxes. Initially, the system recognized all object as ``unknown''; (\textit{b}) a human user teaches all table-top objects to the robot, then, the robot conceptualizes all categories and recognizes all objects correctly; (\textit{c}) afterwards, the user instructs the robot to perform  ``\textit{serve a beer}'' task. The beer object is detected and grasped by the robot. (\textit{d}) The robot moves the beer object over the mug object and serve the drink.}
\label{fig:figure3}
\end{figure*}

\begin{wrapfigure}{r}{0.3\linewidth}
  \vspace{-5mm}
  \begin{center}
    \includegraphics[width=\linewidth]{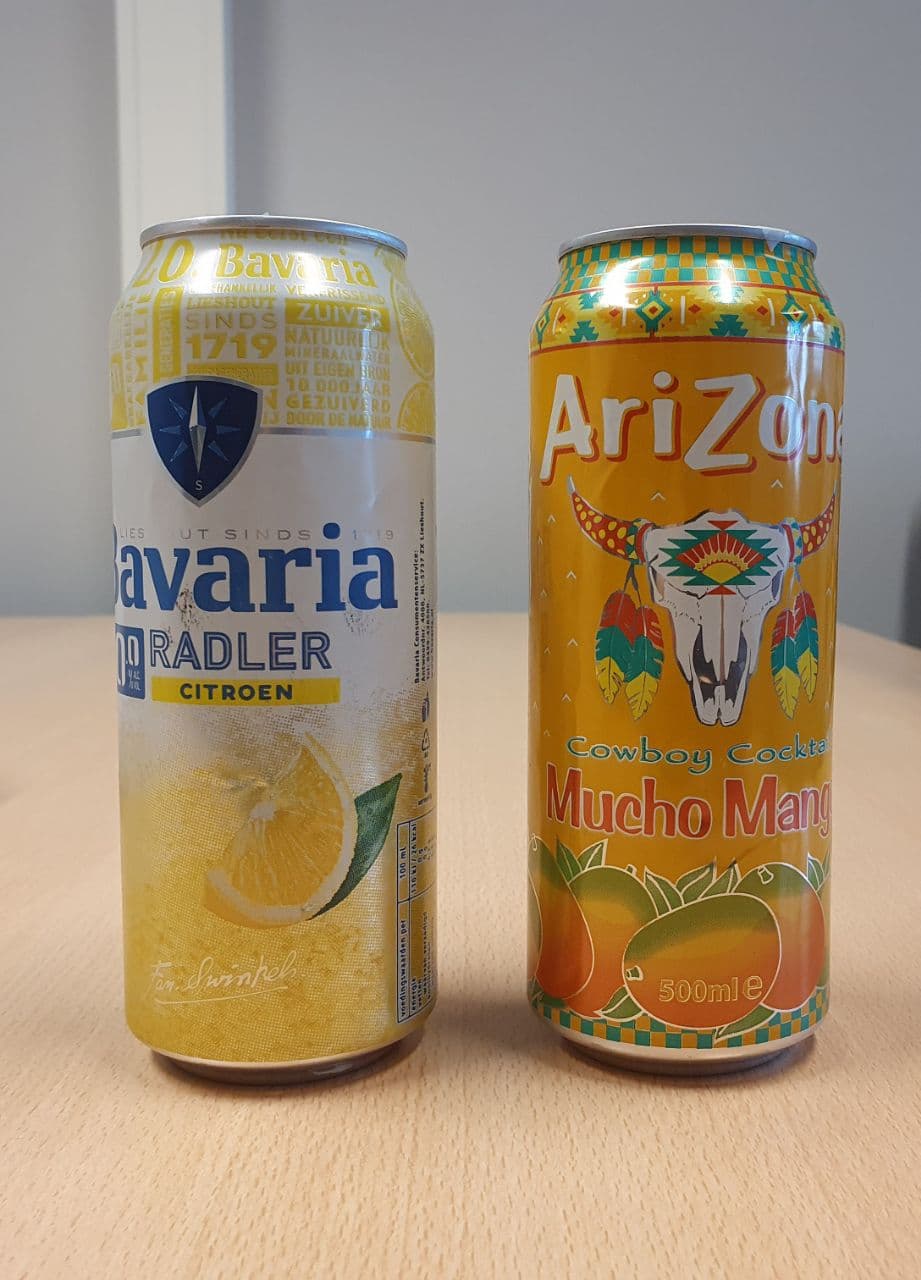}
  \end{center}
  \caption{\small Two very similar objects used in this experiment.}
  \label{fig:similarity}
  \vspace{-1mm}
\end{wrapfigure}
To show the real-time performance of the proposed approach, we performed a real-robot experiment in the context of serving a drink. Our experimental setup is shown Fig.~\ref{fig:setup}. The setup consists of a table with five different objects, namely a \textit{BeerCan}, \textit{CocktailCan}, \textit{Mug}, \textit{Oreo}, and \textit{Vase}. A Kinect sensor is used as the perception device and a UR5e robot arm is employed as the active device. In this experiment, the proposed approach is integrated into the RACE cognitive robotic systems~\cite{oliveira20163d} as a learning and recognition module. Note that, we used three similar objects (i.e., Oreo, Mug, and Vase objects) and two very similar objects (the two types of cans as shown in Fig.~\ref{fig:similarity}) to assess the performance of the proposed approach in terms of open-ended learning and recognition of basic and fine-grained objects using very few examples. 
Fig.~\ref{fig:figure3} shows snapshots of the demonstration. 

In this experiment, the robot should detect and recognize all table-top objects, as indicated by the green bounding boxes and red texts on top of the objects. Initially, the robot does not have any information about the objects, therefore, all objects are recognized as \textit{unknown} category. A human user is involved in the learning loop, The user can interact with the robot using a provided graphical menu to teach the robot new object categories and also provide corrective feedback whenever it is necessary (e.g., misclassifications happen). This way the robot is able to obtain the new data-points from the users in an interactive way. The user starts providing the robot with the respective category labels. As more categories are introduced, the robot learns and recognizes each of them as different categories and not as other instances of previously introduced categories. This demonstration showed that our approach is able to learn about object categories in an interactive and open-ended fashion. Furthermore, it shows the descriptive power of the proposed approach by learning and recognizing very similar objects using very few examples on-site. When the command to ``\textit{serve a beer}'' is given by the user, the robot retrieves the pose of the \textit{BeerCan} object, goes into the pre-grasp pose of the object, and finally grasps and picks it up. the robot then moves the grasped \textit{BeerCan} over the \textit{Mug}, and serve the drink. With this real-time robot demonstration, it has been shown that the robot is able to recognize objects from different orientations, learn new categories in an open-ended fashion, and perform object recognition in real-time ($\approx 30$Hz loop rate). A video of this demonstration is available at: \href{https://youtu.be/dB5s5x6m6WY}{\cblue{https://youtu.be/dB5s5x6m6WY}}.


\section{Conclusions}
\label{sec:conclusion}

In this work, we proposed a deep learning-based approach for 3D object recognition, which enables robots to learn about new object categories in an interactive and open-ended fashion. We encoded both shape and texture information to produce a global scale- and pose-invariant object descriptor. We showed that the obtained representation is descriptive enough to represent both basic and fine-grained object categories. In particular, we rendered rotation and scale-invariant (depth) orthographic projections of an object to encode the shape feature, and from the view that has maximum entropy, we rendered a color image to represent the texture of the object. Afterward, shape and texture projections were fed separately to two networks: one to encode the shape (i.e., we mainly used MobileNet), and the other one was used for encoding the texture information (i.e., we evaluated both MobileNet and DenseNet for this purpose). Afterward, the obtained feature vectors from the shape encoding stream and color encoding stream were combined in a weighted feature vector. The obtained object description was finally used for classification purposes. The proposed approach was analyzed in an offline setting and online experiment, where a simulated human teacher was involved in the learning loop. Experimental results showed that the proposed network architecture outperformed the selected state-of-the-art approaches in terms of object classification accuracy and scalability. Furthermore, we perform a real robot experiment in the context of \textit{serve\_a\_beer} scenario to show the real-time performance of the proposed approach. In the case of offline evaluation, DenseNet proved superior over the MobileNet both in terms of descriptiveness and computational time. Interestingly, throughout online evaluation, DenseNet showed slightly lower descriptiveness than MobileNet. For future work, we plan to investigate how the rendered orthographic projections can be used for object grasping purposes, and then, develop a deep learning architecture to do simultaneous object recognition and grasping.



\begin{thebibliography}{00}
\bibitem{orthographicNET}
Hamidreza Kasaei.
\newblock Orthographicnet: A deep learning approach for 3d object recognition
  in open-ended domains.
\newblock In {\em IEEE/ASME Transactions on Mechatronics}, 2020.

\bibitem{investigating}
Hamidreza Kasaei, Maryam Ghorbani, Jits Schilperoort, and Wessel Rest.
\newblock Investigating the importance of shape features, color constancy,
  color spaces and similarity measures in open-ended 3d object recognition.
\newblock In {\em Intelligent Service Robotics 14}, 2021.

\bibitem{ColorNET}
Shreyank Gowda and Chun Yuan.
\newblock In {\em ColorNet: Investigating the Importance of Color Spaces for Image
  Classification}, pages 581--596.
\newblock Lecture Notes in Computer Science, vol 11364. Springer, 2019.

\bibitem{al2017review}
Ahmed Ali~Mohammed Al-Saffar, Hai Tao, and Mohammed~Ahmed Talab.
\newblock Review of deep convolution neural network in image classification.
\newblock In {\em 2017 International Conference on Radar, Antenna, Microwave,
  Electronics, and Telecommunications (ICRAMET)}, pages 26--31. IEEE, 2017.

\bibitem{simonyan2014very}
Karen Simonyan and Andrew Zisserman.
\newblock Very deep convolutional networks for large-scale image recognition.
\newblock In {\em 3rd IAPR Asian Conference on Pattern Recognition}, 2015.

\bibitem{zhang2019deeper}
Zhipeng Zhang and Houwen Peng.
\newblock Deeper and wider siamese networks for real-time visual tracking.
\newblock In {\em Proceedings of the IEEE Conference on Computer Vision and
  Pattern Recognition}, pages 4591--4600, 2019.

\bibitem{cheng2017survey}
Yu~Cheng, Duo Wang, Pan Zhou, and Tao Zhang.
\newblock A survey of model compression and acceleration for deep neural
  networks.
\newblock In {\em Artificial Intelligence Review 53}, 5113–5155, 2020.

\bibitem{lucas1995towards}
SM~Lucas.
\newblock Towards the open ended evolution of neural networks.
\newblock In {\em First International Conference on Genetic Algorithms in Engineering Systems: Innovations and Applications}, Sheffield, UK, pp. 388-393, 1995.

\bibitem{kasaei2015interactive}
S~Hamidreza Kasaei, Miguel Oliveira, Gi~Hyun Lim, Lu{\'\i}s~Seabra Lopes, and
  Ana~Maria Tom{\'e}.
\newblock Interactive open-ended learning for 3d object recognition: An
  approach and experiments.
\newblock In {\em Journal of Intelligent and Robotic Systems}, 80(3-4):537--553,
  2015.

\bibitem{oliveira2015concurrent}
Miguel Oliveira, Lu{\'\i}s~Seabra Lopes, Gi~Hyun Lim, S~Hamidreza Kasaei,
  Angel~D Sappa, and Ana~Maria Tom{\'e}.
\newblock Concurrent learning of visual codebooks and object categories in
  open-ended domains.
\newblock In {\em 2015 IEEE/RSJ International Conference on Intelligent Robots
  and Systems (IROS)}, pages 2488--2495. IEEE, 2015.

\bibitem{imagenet}
Jia Deng, Wei Dong, Richard Socher, Li-Jia Li, Kai Li, and Fei~Fei Li.
\newblock Imagenet: a large-scale hierarchical image database.
\newblock In {\em IEEE Conference on Computer Vision and Pattern Recognition}, Miami, FL, pp. 248-255, 2009.

\bibitem{view1}
T. Yu, J. Meng and J. Yuan.
\newblock Multi-view Harmonized Bilinear Network for 3D Object Recognition.
\newblock In {\em IEEE/CVF Conference on Computer Vision and Pattern Recognition, Salt Lake City, UT,} pp. 186-194. 2018.

\bibitem{kanezaki2018rotationnet}
Asako Kanezaki, Yasuyuki Matsushita, and Yoshifumi Nishida.
\newblock Rotationnet: Joint object categorization and pose estimation using
  multiviews from unsupervised viewpoints.
\newblock In {\em Proceedings of the IEEE Conference on Computer Vision and
  Pattern Recognition}, pages 5010--5019, 2018.
  
\bibitem{mvcnn}
Su, H., Maji, S., Kalogerakis, E., and Learned-Miller, E.
\newblock Multi-view convolutional neural networks for 3d shape recognition.
\newblock In {\em Proceedings of the IEEE international conference on computer vision}, pp. 945-953. 2015.

\bibitem{volume1}
Qi, C. R., Su, H., Nießner, M., Dai, A., Yan, M., and Guibas, L. J. 
\newblock Volumetric and multi-view cnns for object classification on 3d data.
\newblock In {\em Proceedings of the IEEE conference on computer vision and pattern recognition}. 2016.

\bibitem{volume2}
Wu, J., Zhang, C., Xue, T., Freeman, B., and Tenenbaum, J.
\newblock Learning a probabilistic latent space of object shapes via 3d generative-adversarial modeling. 
\newblock {\em Advances in neural information processing systems}, 29, 82-90. 2016.
  
\bibitem{pointnet}
Qi, C. R., Su, H., Mo, K., and Guibas, L. J.
\newblock Pointnet: Deep learning on point sets for 3d classification and segmentation.
\newblock In {\em Proceedings of the IEEE conference on computer vision and pattern recognition}. 2017.

\bibitem{pointnet++}
Qi, C. R., Yi, L., Su, H., and Guibas, L. J. 
\newblock Pointnet++: Deep hierarchical feature learning on point sets in a metric space. 
\newblock In {\em Advances in neural information processing systems}, 5099-5108. 2017.

\bibitem{Ayoobi2020LocalHDPI}
H.~Ayoobi, H.~Kasaei, M.~Cao, Rineke Verbrugge, and B.~Verheij.
\newblock Local-hdp : Interactive open-ended 3d object categorization.
\newblock In {\em ECCV 2020}, 2020.

\bibitem{belongie2002shape}
Serge Belongie, Jitendra Malik, and Jan Puzicha.
\newblock Shape matching and object recognition using shape contexts.
\newblock {\em IEEE transactions on pattern analysis and machine intelligence},
  24(4):509--522, 2002.

\bibitem{kazmi2013survey}
Ismail~Khalid Kazmi, Lihua You, and Jian~Jun Zhang.
\newblock A survey of 2d and 3d shape descriptors.
\newblock In {\em 2013 10th International Conference Computer Graphics, Imaging
  and Visualization}, pages 1--10. IEEE, 2013.

\bibitem{good}
Hamidreza Kasaei, Ana Tomé, Luís Seabra~Lopes, and Miguel Oliveira.
\newblock Good: A global orthographic object descriptor for 3d object
  recognition and manipulation.
\newblock In {\em Pattern Recognition Letters}, 83, 07 2016.

\bibitem{krizhevsky2012imagenet}
Alex Krizhevsky, Ilya Sutskever, and Geoffrey~E Hinton.
\newblock Imagenet classification with deep convolutional neural networks.
\newblock In {\em Advances in neural information processing systems}, pages
  1097--1105, 2012.

\bibitem{color2}
J. Krause, M. Stark, J. Deng and L. Fei-Fei.
\newblock 3D Object Representations for Fine-Grained Categorization.
\newblock In {\em IEEE International Conference on Computer Vision Workshops, Sydney, NSW,} pp. 554-561. 2013.

\bibitem{poth2016breaking}
Christian~H Poth and Werner~X Schneider.
\newblock Breaking object correspondence across saccades impairs object
  recognition: The role of color and luminance.
\newblock In {\em Journal of Vision}, 16(11):1--1, 2016.

\bibitem{geirhos2018imagenet}
Robert Geirhos, Patricia Rubisch, Claudio Michaelis, Matthias Bethge, Felix~A
  Wichmann, and Wieland Brendel.
\newblock Imagenet-trained cnns are biased towards texture; increasing shape
  bias improves accuracy and robustness.
\newblock In {\em arXiv preprint arXiv:1811.12231}, 2018.

\bibitem{ColorOR}
Inês Bramão, Alexandra Reis, Karl~Magnus Petersson, and Luís Faísca.
\newblock The role of color information on object recognition: A review and
  meta-analysis.
\newblock In {\em Acta psychologica}, 138:244--53, 07 2011.

\bibitem{tsai2018simultaneous}
Chi-Yi Tsai and Shu-Hsiang Tsai.
\newblock Simultaneous 3d object recognition and pose estimation based on rgb-d
  images.
\newblock In {\em IEEE Access}, 6:28859--28869, 2018.

\bibitem{zia2017rgb}
Saman Zia, Buket Yuksel, Deniz Yuret, and Yucel Yemez.
\newblock Rgb-d object recognition using deep convolutional neural networks.
\newblock In {\em Proceedings of the IEEE International Conference on Computer
  Vision Workshops}, pages 896--903, 2017.

\bibitem{asif2017rgb}
Umar Asif, Mohammed Bennamoun, and Ferdous~A Sohel.
\newblock Rgb-d object recognition and grasp detection using hierarchical
  cascaded forests.
\newblock In {\em IEEE Transactions on Robotics}, 33(3):547--564, 2017.

\bibitem{cruz2013object}
Jerome Paul~N Cruz, Ma~Lourdes Dimaala, Laurene Gaile~L Francisco, Erica
  Joanna~S Franco, Argel~A Bandala, and Elmer~P Dadios.
\newblock Object recognition and detection by shape and color pattern
  recognition utilizing artificial neural networks.
\newblock In {\em 2013 International Conference of Information and
  Communication Technology (ICoICT)}, pages 140--144. IEEE, 2013.

\bibitem{sandler2018mobilenetv2}
Mark Sandler, Andrew Howard, Menglong Zhu, Andrey Zhmoginov, and Liang-Chieh
  Chen.
\newblock Mobilenetv2: Inverted residuals and linear bottlenecks.
\newblock In {\em Proceedings of the IEEE conference on computer vision and
  pattern recognition}, pages 4510--4520, 2018.

\bibitem{huang2017densely}
Gao Huang, Zhuang Liu, Laurens Van Der~Maaten, and Kilian~Q Weinberger.
\newblock Densely connected convolutional networks.
\newblock In {\em Proceedings of the IEEE conference on computer vision and
  pattern recognition}, pages 4700--4708, 2017.

\bibitem{washington-d}
Kevin Lai, Liefeng Bo, Xiaofeng Ren, and Dieter Fox.
\newblock A large-scale hierarchical multi-view rgb-d object dataset.
\newblock In {\em IEEE International Conference on Robotics and Automation}, Shanghai, pp. 1817-1824, 2011.

\bibitem{SGD}
Bottou, L. 
\newblock Large-scale machine learning with stochastic gradient descent. 
\newblock In {\em Proceedings of COMPSTAT'2010,} pp. 177-186. 2010.

\bibitem{oliveira20163d}
Miguel Oliveira, Lu{\'\i}s~Seabra Lopes, Gi~Hyun Lim, S~Hamidreza Kasaei,
  Ana~Maria Tom{\'e}, and Aneesh Chauhan.
\newblock 3d object perception and perceptual learning in the race project.
\newblock In {\em Robotics and Autonomous Systems}, 75:614--626, 2016.

\bibitem{bow}
Kasaei Hamidreza, Oliveira Miguel, Gi Hyun Lim, Luís Seabra Lopes, Ana Maria Tomé.
\newblock Towards lifelong assistive robotics: A tight coupling between object perception and manipulation 
\newblock In {\em Neurocomputing, vol. 291}, pp. 151–166, 2018.

\bibitem{local-lda}
S. H. Kasaei, L. S. Lopes and A. M. Tomé.
\newblock Local-LDA: Open-Ended Learning of Latent Topics for 3D Object Recognition.
\newblock In {\em IEEE Transactions on Pattern Analysis and Machine Intelligence}, vol. 42, no. 10, pp. 2567-2580, 2020.

\bibitem{liu2019soft}
Liu, Shichen and Li, Tianye and Chen, Weikai and Li, Hao
\newblock Soft rasterizer: A differentiable renderer for image-based 3D reasoning
\newblock In {\em Proceedings of the IEEE International Conference on Computer Vision}, pages 7708-7717, 2019.

\end{thebibliography}
\end{document}